\documentclass[times,referee,twocolumn,final,authoryear]{elsarticle}

\usepackage{ycviu}
\usepackage{framed,multirow}

\usepackage{graphicx}
\usepackage{amsmath,amssymb} %
\usepackage{latexsym}

\usepackage{url}
\usepackage[dvipsnames]{xcolor}
\definecolor{newcolor}{rgb}{.8,.349,.1}
\usepackage{subcaption,rotating,multirow}
\usepackage{sidecap}
\usepackage{pdflscape}

\usepackage[normalem]{ulem}
\usepackage{xifthen}
\usepackage{tabularx}

\usepackage{array}

\usepackage[hidelinks]{hyperref}

\usepackage{xspace}
\makeatletter
\DeclareRobustCommand\onedot{\futurelet\@let@token\@onedot}
\def\@onedot{\ifx\@let@token.\else.\null\fi\xspace}

\def\eg{\emph{e.g}\onedot} 
\def\ie{\emph{i.e}\onedot} 
\def\cf{\emph{c.f}\onedot}

\makeatother

\newcommand{\myparagraph}[1]{\paragraph{\textbf{#1}}}

\let\wip\emph

\def\case#1#2{\def\mytemp{#1}\ifx\mytemp\isswitch\def\isdefault{1}#2\fi\ignorespaces}
\def\default#1{\if0\isdefault#1\fi\ignorespaces}
\def\isdefault{0}
\def\isswitch{0}

\protected\def\wip#1{\ignorespaces\def\isdefault{0}\def\isswitch{#1}\ignorespaces
	\case{L1}{$L_1$}
	\case{L1M}{$L_1^{\textrm{M}}$}
    \case{ig}{$\text{IG}$} %
    \case{igu}{$\text{IG}_{\textrm{U}}$} %
    \case{igs}{$\text{IG}_{\textrm{S}}$} %
	\case{ig6}{$\text{IG6}$} %
    \case{ig6u}{$\text{IG6-U}$} %
    \case{ig6s}{$\text{IG6-S}$} %
    \case{ig6m}{$\text{IG6-M}$} %
    \case{ig6mu}{$\text{IG6-MU}$} %
    \case{ig6ms}{$\text{IG6-MS}$} %
    \case{iga}{$\text{IGA}$}
    \case{igk}{$\text{IGS}$}    
    \case{igam}{$\text{IGA-M}$}
    \case{igkm}{$\text{IGS-M}$}    
    \case{igamu}{$\text{IGA-MU}$}
    \case{igkmu}{$\text{IGS-MU}$}    
    \case{igsu}{$\text{IGS-M}$}
    \case{p2p}{P2P} %
    \case{p30}{P2P-30} %
    \case{p65}{P2P-6$\times$5} %
    \case{pm30}{P2P-M30} %
    \default{\emph{#1}}
    \xspace
}
\def\>{$\rightarrow$}
\def\<{$\Leftarrow$}

\DeclareMathOperator{\LL}{\mathcal{L}}

\newcommand{\Dis}{\mathcal{D}}
\newcommand{\Gen}{\mathcal{G}}

\graphicspath{{./figures/}}

\journal{Computer Vision and Image Understanding}
\begin{document}

\ifpreprint
  \setcounter{page}{1}
\else
  \setcounter{page}{1}
\fi

\begin{frontmatter}

\title{IterGANs: Iterative GANs to Learn and Control 3D Object Transformation}

\author[1]{Ysbrand \snm{Galama}\corref{uva}}%
	\ead{ysbrand.galama@tomtom.com}
\author[2]{Thomas \snm{Mensink}\corref{uva}}%
	\ead{mensink@google.com}

\address[1]{TomTom, Amsterdam, The Netherlands}
\address[2]{Google Research, Amsterdam, The Netherlands}

\cortext[uva]{The main body of this research was performed while both authors where with the Computer Vision group, University of Amsterdam, the Netherlands.}

\received{13 July 2018}
\finalform{10 May 2013}
\accepted{13 May 2013}
\availableonline{15 May 2013}
\communicated{S. Sarkar}

\begin{abstract}
We are interested in learning visual representations which allow for 3D manipulations of visual objects based on a single 2D image.
We cast this into an image-to-image transformation task, and propose Iterative Generative Adversarial Networks (IterGANs) which iteratively transform an input image into an output image. Our models learn a visual representation that can be used for objects seen in training, but also for never seen objects.
Since object manipulation requires a full understanding of the geometry and appearance of the object, our IterGANs learn an implicit 3D model and a full appearance model of the object, which are both inferred from a single (test) image.
Two advantages of IterGANs are that the intermediate generated images can be used for an additional supervision signal, even in an unsupervised fashion, and that the number of iterations can be used as a control signal to steer the transformation. 
Experiments on rotated objects and scenes show how IterGANs help with the generation process.
\end{abstract}


\end{frontmatter}

\section{Introduction}
In this paper we are interested in manipulating visual objects and scenes, without resorting to external provided (CAD) models or advanced (3D/depth) sensing techniques.
To be more specific, we focus on rotating objects and rotating the camera viewpoint of scene from a single 2D image.
Manipulating objects require an expectation about the appearance and the geometrical structure of the \emph{unseen} part of the object. 
Humans clearly have such an expectation based on an understanding of the physics of the world, the continuity of objects, and previously seen (related) objects and scenes.
We aim to learn such an 3D understanding, inferred from single 2D images.

\begin{figure}[h]	
	\centering
	{\includegraphics[height=28mm]{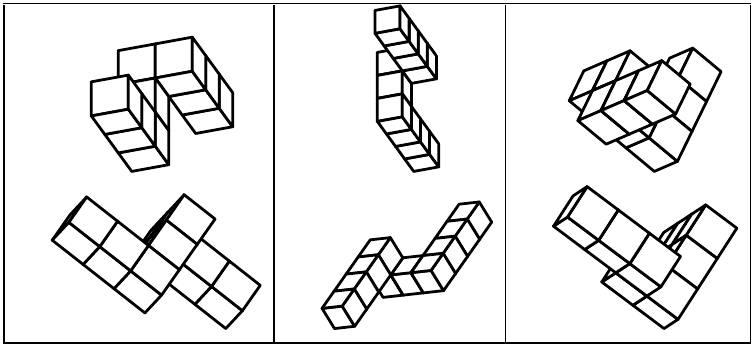}}\vspace{-3mm}
	{\centerline {\footnotesize \emph{ How long does it take you to find the non-matching pair?}}}\vspace{-1mm}
    \caption{    
    \citet{shepard1971mental} demonstrated the time needed to identify matching pairs of objects depends on the degree of rotation, more rotated objects take longer to identify.
    We introduce IterGANs, which \emph{iteratively} rotate objects for a few degrees each time to reach a target rotation.
    }\vspace{-2mm}
	\label{fig:mental}
\end{figure}

In order to learn a representation for object manipulation, we cast this problem into an image-to-image transformation task, with the goal to transform an input image following a given 3D transformation to an target image.
For this kind of object manipulation, often either stereoscopic cameras~\citep{ko20072d,bruno20103d} or temporal data streams~\citep{pollefeys2008detailed,gibson2003interactive} have been used to infer depth cues, while our aim is to obtain the target image from a single input image only. 
Similarly as humans are able to do so with one eye closed~\citep{vishwanath2013seeing}, there has also been works that aim to reconstruct 3D from a single image~\citep{saxena2009make3d,rematas16pami}, however these typically require external provided 3D object models, \eg balloon shapes~\citep{vicente2013balloon}, or focus on a single class of objects only~\citep{park17cvpr}. 
Our aim, on the other hand, is to learn a general transformation model, which can transform many classes of objects, even objects never seen at train time, based on the fact that appearance and geometrical continuity are (mostly) not object specific but generally applicable.

In this paper, we focus on a specific instance of object manipulation: the object in the target image is a fixed rotation of the input image.
For this task, we propose the use of Iterative Generative Adversarial Networks (IterGANs).
GANs have been used for many (image) generation tasks~\citep{reed2016generative,denton2015deep,radford2015unsupervised}, including image-to-image prediction~\citep{pix2pix2016,Zhu_2017_ICCV}. 
Our proposed IterGANs are a special kind of GANs, where the input image is fed to the generator, and the output of the generator is iteratively fed back into the generator for a fixed number of iterations to generate the output image.
The number of iterations is either predefined, or used as a control mechanism to steer the degree of image rotation. 

The underlying hypothesis of the IterGAN is, it is easier to rotate an object for a few degrees, than for a large rotation. 
This has experimentally been shown by~\citet{shepard1971mental}, by measuring the reaction time of humans to identify whether two rotated objects are the same.
The study shows that there exists a linear dependence between the reaction time and the degree of rotation between the two objects, see~\autoref{fig:mental}. 

The iterative nature of IterGANs has two particular advantages over a single image-to-image GAN for image manipulation.
First, IterGANs break long range dependencies between the pixels of the input image and the pixels of the output image.
A fundamental difference between image manipulation and the image-to-image tasks explored in~\citep{pix2pix2016} is, that when translating a map into an aerial image there exists a one-to-one pixel relation between the input and the output image. In the case of object manipulation, however, pixels have long range dependencies, depending on the geometry and the appearance of the object and the required degree of rotation.
IterGANs break these long dependencies into a series of shorter dependencies. 
Second, IterGANs allow to use intermediate loss functions measuring the quality of the series of intermediate generated images to improve the overall transformation quality.

This paper is an extended version of our ICLR Workshop paper~\citep{galama18openreview}. %
This extended version, includes an supervised intermediate loss function, a stepwise learning approach, and extensive experimental results, on both the ALOI dataset~\citep{geusebroek2005amsterdam} for object rotation, and the VKITTI dataset~\citep{Gaidon:Virtual:CVPR2016} for camera viewpoint scene rotation. 
Our paper is organised as follows. 
Next we will discuss some of the most related work in image-to-image object manipulation.
In \autoref{sec:mm-models} we introduce IterGANs and propose extended loss functions on the intermediate generated images.
We show extensive experimental results in \autoref{sec:ex}, on the ALOI dataset~\citep{geusebroek2005amsterdam} for object rotation, and the VKITTI dataset~\citep{Gaidon:Virtual:CVPR2016} for camera viewpoint scene rotation.
Finally, we conclude the paper in \autoref{sec:conc}.
\section{Related work}
There is vast amount of related work in the field of 3D reconstruction and image generation.
Here, we only highlight the most relevant methods with respect to our proposed models.
For a more detailed overview see for 3D reconstruction~\citep{li2015overview} and image generation~\citep{Zhu_2017_ICCV}.

\begin{figure*}[t]
	\centering
	\resizebox{.9\textwidth}{!}{
	\subcaptionbox{IterGAN base\label{fig:itergan}}{\includegraphics[width=.3\textwidth]{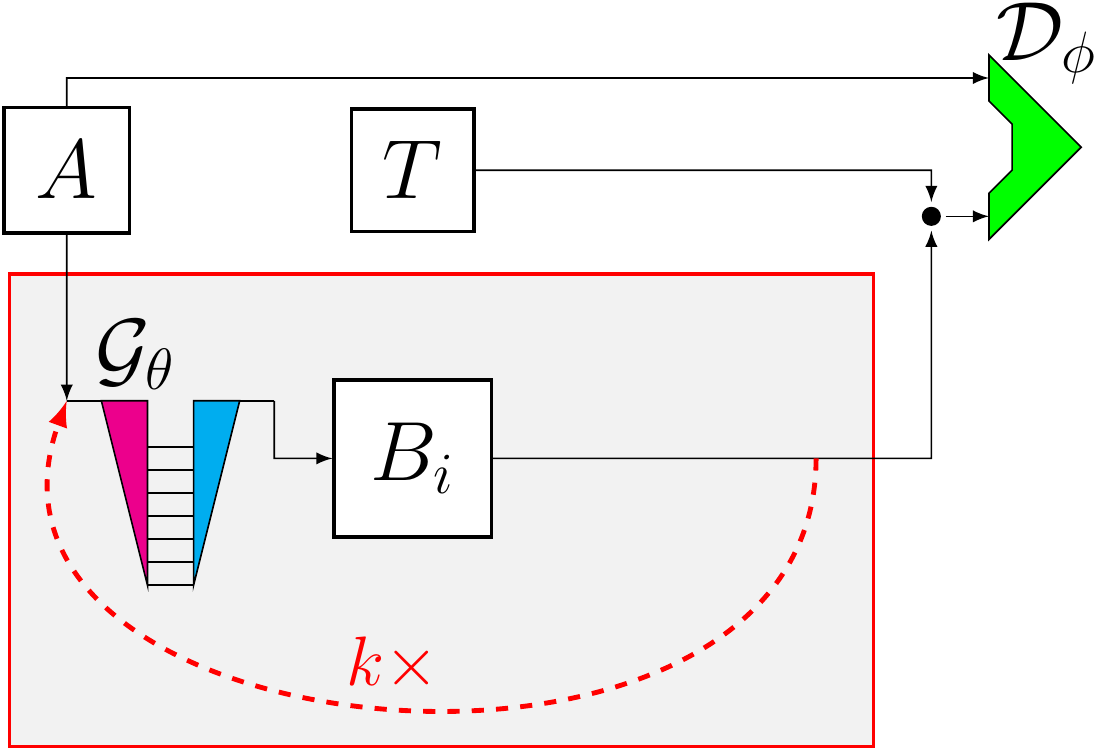}}
	\quad
	\subcaptionbox{Unsupervised IDL\label{fig:unsupervised}}{\includegraphics[width=.3\textwidth]{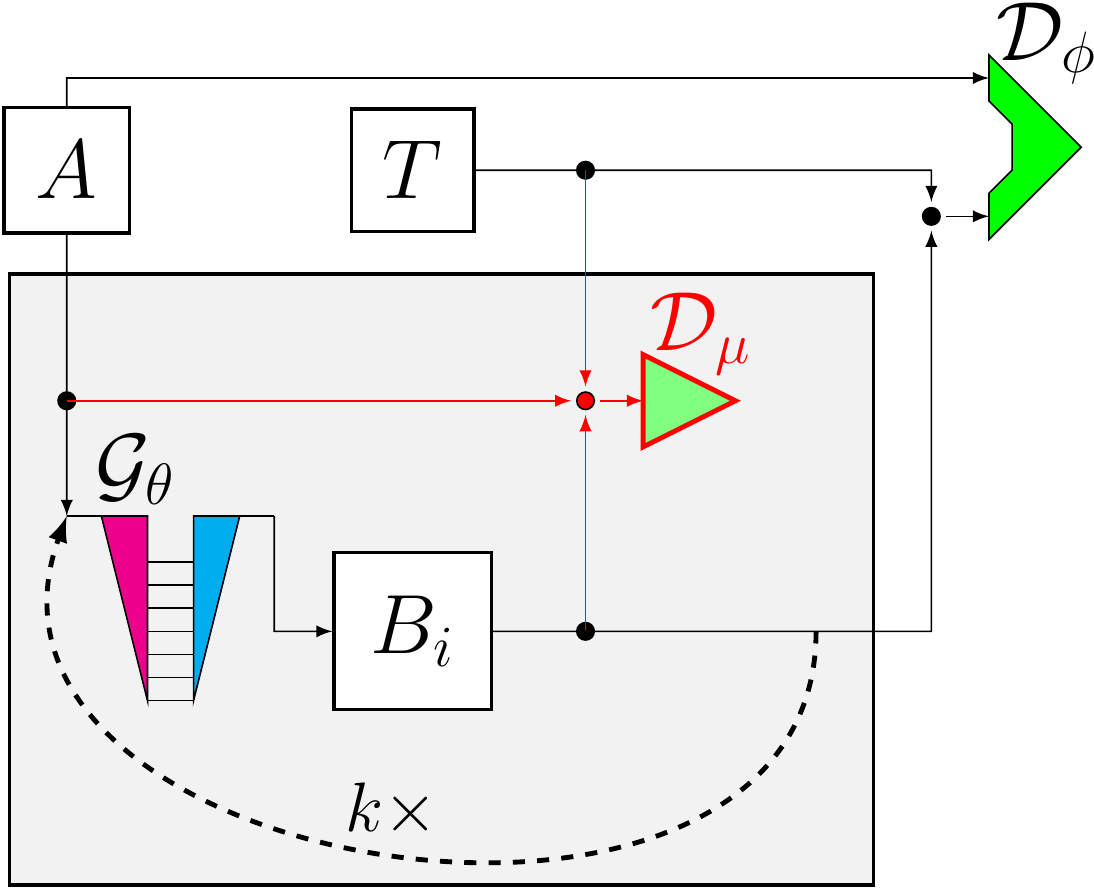}}
	\quad
	\subcaptionbox{Supervised IDL\label{fig:supervised}}{\includegraphics[width=.3\textwidth]{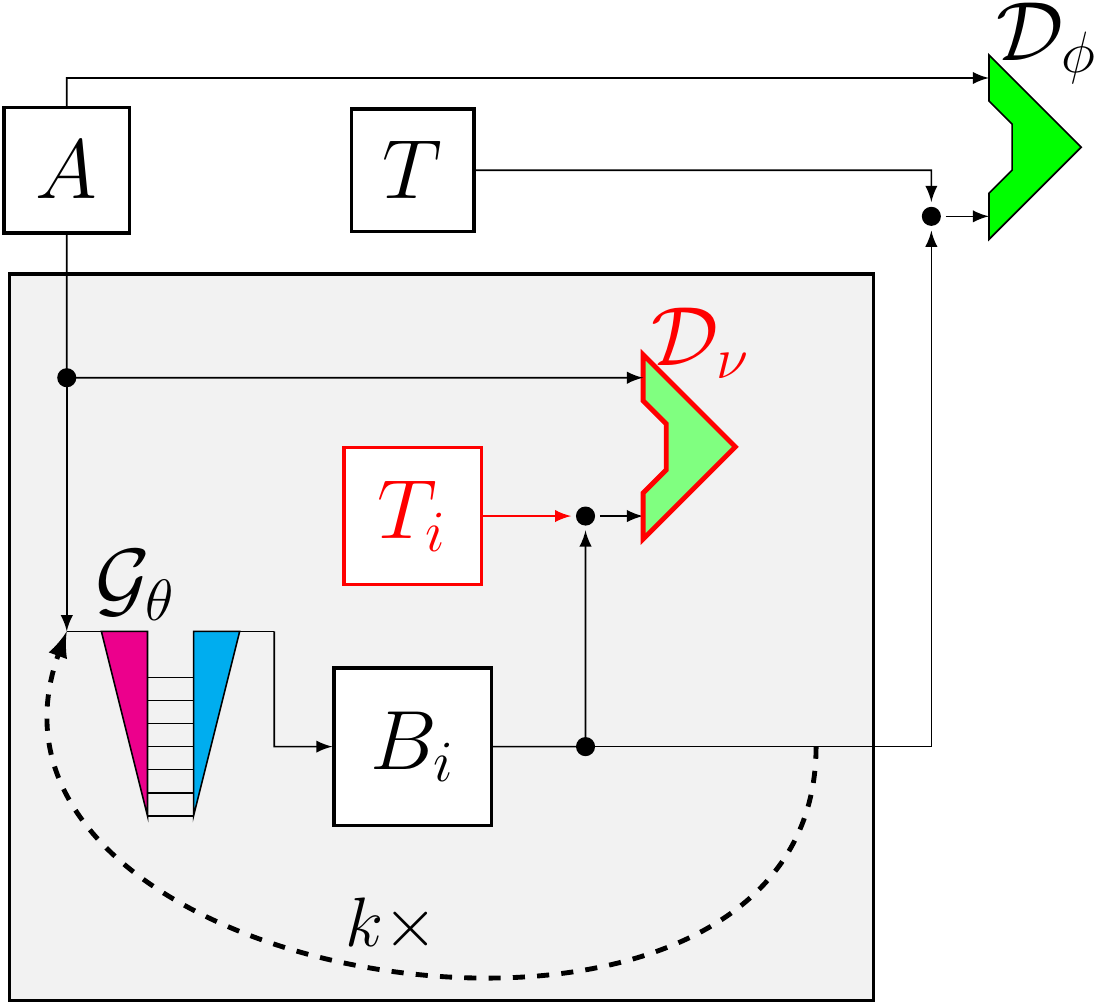}}
	}
	\caption{IterGAN framework: the iterative nature of IterGANs (\emph{left}) allows for additional discriminators on the intermediate generated images to steer the learning.
	In this paper we introduce two types of intermediate discriminator loss functions (IDL).
	\textbf{Unsupervised IDL} (\emph{middle}) aims to tell apart generated intermediate images ($B^i$) from real images ($A$ or $T$). 
	\textbf{Supervised IDL} (\emph{right}) aims to discriminate input-generated $(A,B^i)$ image pairs from the real pairs $(A,T^i$).
	}
\end{figure*}

\myparagraph{3D reconstruction from 2D images}
There are several techniques for generating 3D models from 2D data, differing in both the type of data, and the type of environment.
It is possible to create a point-cloud from video using Structure from Motion~\citep{koenderink1991affine,nister2005preemptive}, or to fit or deform polygonal objects to images~\citep{rematas16pami,suveg2004reconstruction,vicente2013balloon}.

In recent years, there have also been attempts to use deep learning.
These techniques also allow forgoing 3D models, thus using only 2D images to describe the 3D environment.
Such an approach has been used to classify objects~\citep{su2015multi}, generate different viewpoints from descriptors~\citep{dosovitskiy2015learning} or creating the frames for 3D movies~\citep{xie2016deep3d}.

In contrast to their work, we focus on changing the orientation of an object in the image or the viewpoint of a scene in a given image.
Therefore our models not only need to construct the view~\citep{dosovitskiy2015learning}, but also perceive an input image, and capture more than only a disparity map, \eg~\citep{xie2016deep3d}.
Since our desired output is an image, we use image-to-image GANs to transform the image.

\myparagraph{Image manipulation with GANs}
Generative Adversarial Networks (GANs, ~\citet{goodfellow2014generative}) have been shown to be successful for generating visually pleasing images, \eg~\citep{reed2016generative,denton2015deep,radford2015unsupervised}.
For image-to-image translation, where the goal is to translate an input images (\eg sketch) to an output image (\eg photo), conditional GANs~\citep{mirza2014conditional} have been used by~\citep{pix2pix2016} in their Pix2Pix paper, where the input image is the conditional.
The Pix2Pix paradigm has sparked many image manipulation tasks into image-to-image translation, including aging a face from a single image~\citep{antipov2017face}, or provide it with glasses and a shave~\citep{shen2017learning}, or change the main object of an image, \eg~a cat to a dog~\citep{liang18eccv}.
We also use the Pix2Pix image-to-image GANs, but apply and extend it for object manipulation, to generate images of rotated objects/scenes.

Since GANs are notoriously difficult to train, in part due to mode collapse, many variants have been introduced, focusing on the optimisation and the loss functions of the network, \eg~\citep{salimans16nips,arjovsky17icml}.
Another line of research is to use the network architecture to guide the learning process, for example by imposing a cyclic or dual GAN architecture~\citep{zhu17iccv,yi17iccv}, which exploits the insight that if image $A$ transforms into image $\hat{B}$, then with an inverse transformation image $\hat{B}$ should transform back into $A$.
The same insight of cyclic generation is also used for left-right consistency when depth is generated from a single image, trained on paired left-right images without depth groundtruth~\citep{pilzer18eccv}.
Instead of using two separate generators, one for the transformation and for the inverse transformation, the cyclic reasoning could also be used with a single generator with some control vector ~\citep{liang18eccv,choi18cvpr}.
We also change the GAN architecture, where we explicitly let the GAN transform the object iteratively, and instead of using a control vector, we rather use the number of iterations of the generator as a control function.

Finally, an important line of research is on improving the quality of the synthesised images. For example by progressively increasing the depth of the generator and the discriminator over the coarse of training~\citep{karras18iclr} improves the quality and stability of training in higher resolutions. In~\citep{wang18cvpr} a coarse-to-fine generator is introduced, where first a small image is generated, which is then used as conditional to generate a larger output image.
This two step zooming process into the details of a high resolution image, is similar in how our IterGAN model iteratively rotates an object to a final rotation.

\myparagraph{Novel viewpoint estimation}
A more specific form of image manipulation is \emph{Novel viewpoint estimation}, which has a slightly different goal than 3D reconstruction, since the output is again a 2D image.
A new viewpoint can be estimated using voxel projections~\citep{yan16nips}, or GANs to estimate the frontal view of a face~\citep{huang2017beyond}, or transforming a single image with viewpoint estimation~\citep{zhou16eccv}.
The quality of the generated image can be improved by using multiple input images~\citep{zhou16eccv}, or using a second network to refine the results of a flow-network~\citep{park17cvpr}. 
Existing works on viewpoint transformation have been conducted to synthesize novel views of the same object using synthesised data from CAD models.
In our paper, we use a heterogeneous dataset of real world images taken under constraint conditions, instead of synthetic images of a single object class.

\section{IterGANs}
\label{sec:mm-models}
Iterative GANs are image-to-image GANs, where the generator is called iteratively for $k$ steps:
\begin{equation}
	B^k = \Gen_\theta(\Gen_\theta(\Gen_\theta(\Gen_\theta(\Gen_\theta(\Gen_\theta(A)))))) = \Gen^k_{\theta}(A),
	\label{eq:gen}
\end{equation}
where the output image $B^k$ is generated by rotating the input image $A$ in $k$ small steps using the same generator $\Gen$.
While this iterative generation can not guarantee a specific degree rotation, \eg $6 \times 5^\circ = 30^\circ$, 
it breaks the long dependencies between pixels of rotated objects, and therefore the generator should be easier to learn.
The number of steps could also be used to control the degree of rotation, by varying the number of iterations, which we explore in \autoref{sec:control}.
The iterative nature of the IterGAN is illustrated in~\autoref{fig:itergan}, we refer to our IterGAN network as \wip{ig}.

The iterative nature of IterGANs resembles recurrent neural network modules, like LSTM~\citep{lstm97} or GRUs~\citep{gru14}.
Typically use cases for RNNs in computer vision include classifying a sequence of inputs, \eg to classify actions in video~\citep{ng15cvpr}, or to produce a sequence conditioned on an input, \eg describe an image with a caption~\citep{vinyals15cvpr}.

In contrast to the first, IterGANs only receive a \emph{single} input, instead of a sequence (sentence or video-frames) and produce a single (image) output.
The difference with the latter is more subtle: both the image captioning as well as our IterGANs are conditioned on a single input.
However, there are two main differences: 
(a) LSTMs propagate both the (sampled) output of the previous layer combined with a hidden state, while the proposed IterGANs only propagate the generated image of the previous layer; 
(b) the quality of the caption is determined by the combination of words, and the quality of an individual word is difficult to measure, while for IterGANs we are mainly interested in the output after iteration $k$ and the quality of each generated image is independently measurable.

\myparagraph{Discriminator and Generator Loss Functions}
We use the same network architectures as in Pix2Pix~\citep{pix2pix2016}.
IterGANs could be seen as an extension of Pix2Pix, and are identical when $k=1$ is used.
Since in IterGANs the same generator is applied iteratively, it also has the same number of parameters as the Pix2Pix models.
For training we use the generator and discriminator losses of~\citet{pix2pix2016}.:
\begin{align}
	\LL_\Gen^{(\text{\wip{ig}})} &= H[\Dis_\phi(A, B^k), 1] + \lambda_{L_1} \ L_1(B^k, T)\label{eq:mm-lgen-m1},\\
	\LL_\Dis^{(\text{\wip{ig}})} &= H[\Dis_\phi(A, T), 1] + H[\Dis_\phi(A, B^k), 0],\label{eq:mm-ldis-m1}
\end{align}
where $A$ denotes the input image, $T$ the target image, and $B^k= \Gen^k_{\theta}(A)$ the generated output image after step $k$.
The generator ($\LL_\Gen$) combines the cross-entropy loss ($H$) for predicting \emph{real} for the image pair $A$, $B^k$, \ie a low loss value is obtained when the discriminator is fooled that $B^k$ is a real image, with the $L_1$ loss between the generated image $B^k$ and the target image $T$.
The discriminator ($\LL_\Dis$) combines the cross-entropy loss for predicting \emph{real} for the input pair $A$ and $T$ and \emph{generated} for the input pair $A$ and $B^k$.

\myparagraph{Object mask specific reconstruction loss}
The L1 loss equally weights each pixel in the image. 
For our object rotation, we would like to focus more on the pixels of the rotated object, rather than the (black) background. 
Therefore, we use a variant of the L1-loss, which uses a binary object mask $M$:
\begin{align}
	L_1^{\textrm{M}} &= \frac{2}{|M|} \sum_{xy} M_{xy} L_{xy} + \frac{1}{|\neg M|} \sum_{xy} \neg M_{xy} L_{xy}, \label{eq:L1m}\\ 
	\textrm{with } L_{xy} &= \sum_c |T_{xyc}-B^k_{xyc}|, \nonumber	
\end{align}
where $M \in \{0,1\}$ assigns pixels to object (1) or background (0), these masks are part of the ALOI dataset~\citep{geusebroek2005amsterdam}.
This variant of the L1-loss weights the object twice as important as the background.
Models trained with this $L_1^{\textrm{M}}$-loss are indicated with an `M'.

\begin{figure}[t]
  \centering  
  \includegraphics[width=.8\columnwidth]{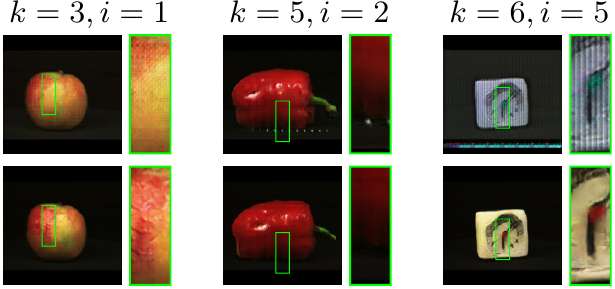}    
  \caption{Examples of generated artefacts in intermediate images (\emph{top row}), which seem independent of specific value for the number of iterations $k = \{3,5,6\}$.
  Interestingly, the artefacts disappear in the final generated image (\emph{bottom row}).
		}  
    \label{fig:artefacts}        
\end{figure}

\subsection{Intermediate Discriminator Loss Functions}
While IterGANs generate intermediate images, the (implicit) assumption that these would be realistic images as well, does not necessarily hold.
In fact, preliminary results reveal that, the iterative generator introduces different types of artefacts in the intermediate generated images.
Interestingly, these are gone in the final generated image, so repeatedly applying the same generator removes the introduced artefacts.
In \autoref{fig:artefacts}, we show examples of intermediate generated images when using $k=\{3,5,6\}$ and each of these have artefacts, \eg switching colours, adding noise or patterns.

In this section we introduce two types of intermediate discriminator loss functions (IDL), to overcome these artefacts and to improve the final image generation quality.
The IDLs aim to guide the learning process to generate realistic intermediate images, by an additional discriminator fed by intermediate generated images.
We propose IDLs which do not require additional target images (\emph{unsupervised}), and which do use additional target images of the intermediate images (\emph{supervised}).
Both models are illustrated in \autoref{fig:unsupervised} and \autoref{fig:supervised}.

\myparagraph{Unsupervised Intermediate Discriminator Loss function}
The unsupervised IDL requires only the (A,T) image pair, already provided to the GAN, yet it includes an additional discriminator, to tell apart image A or T from any of the generated images $\{B^i\}_{i=1}^k$,  unconditioned on the original input image, see \autoref{fig:unsupervised}.
This additional discriminator guides the GAN to generate intermediate images following the distribution of \emph{real} images A and T.
This results in the extended  loss functions:
\begin{align}
	\LL_\Gen^{(\text{\wip{igu}})} &= \LL_\Gen^{(\text{\wip{ig}})} + \lambda_u \ H[\Dis_\mu(B^i), 1]\label{eq:mm-lgen-m3} \\
	\LL_\Dis^{(\text{\wip{igu}})} &= \LL_\Dis^{(\text{\wip{ig}})} + \lambda_u \ \big( H[\Dis_\mu(B^i), 0] + H[\Dis_\mu(A \lor T), 1] \big) \label{eq:mm-ldis-m3}
\end{align}
where an additional cross-entropy loss (H) is added for the generator and the discriminator.
For each image in a batch, we sample a single $B^i$ uniformly from $\{B^i\}_{i=1}^k$, and either $A$ or $T$ is used, $\lambda_u$ is an additional hyper-parameter.
Since no additional labeled data is required, we call this \emph{unsupervised IDL} and indicate models trained with these losses with `U'.

\myparagraph{Supervised Intermediate Discriminator Loss}
The supervised intermediate loss also adds a discriminator, but one conditioned on the input image $A$, and using the intermediate target images $\{T^i\}_{i=1}^k$ for supervision, see \autoref{fig:supervised}.
It aims to discriminate whether the intermediate images are a generated rotation ($\{B^i\}_{i=1}^k$ or a real rotation ($\{T^i\}_{i=1}^k$) from the object depicted in image $A$.
This discriminator is similar to the existing conditional discriminator used by the GAN, yet the difference is that the goal of the main discriminator is to detect if the output is a real $R^\circ$ rotation of the input $A$, while the goal of the added discriminator is to accept an arbitrarily rotation of image $A$.
Supervised IDL results in the following extended losses:
\begin{align}
    \LL_\Gen^{(\text{\wip{igs}})} &= \LL_\Gen^{(\text{\wip{ig}})} + \lambda_s \ H[\Dis_\nu(A,B^i), 1] \label{eq:mm-lgen-m4}\\
    \LL_\Dis^{(\text{\wip{igs}})} &= \LL_\Dis^{(\text{\wip{ig}})} + \lambda_s \left( H[\Dis_\nu(A,B^i), 0] + H[\Dis_\nu(A,T^i), 1] \right)\label{eq:mm-ldis-m4}
\end{align}
where an additional cross-entropy loss (H) is added on the conditional model, and where ($B^i,T^i$) are sampled randomly, with $i= \{1,\ldots,k-1\}$.
Preliminary experiments using also intermediate $L_1^M$-loss between $B^i$ and $T^i$ did not improve the final quality of the generator, and were therefore not further explored.
The proposed losses for the generator and discriminator need the intermediate groundtruth images $\{T^i\}_{i=1}^k$, therefore we call this \emph{supervised IDL}, and indicate models with an `S'.

\subsection{Training IterGANs to Control Object Manipulation}
\label{sec:control}
A notable difference between the unsupervised and supervised IDL is that the implicit versus explicit requirement of the generator to rotate an object for a predefined (fixed) degree of rotation.
The unsupervised model could rotate any degree as long as the final rotation is correct, while the supervised model, is guided to rotate each iteration for the same degree of rotation.
In this section, we follow up on this, by using the number of iterations as a way to control to the desired rotation of the object even more explicit.

The number of iterations could be seen as an explicit control variable, in contrast to learn from implicit control, \eg by adding a control vector to the generator network for a desired output domain/viewpoint, see \eg~\citep{zhou16eccv,park17cvpr,choi18cvpr}. 
Instead of training on input-target pairs with a fixed rotation, we sample a value of $k \in \{1,\ldots,36\}$ and select input-target pairs with corresponding difference of rotation $(5,10,\ldots,180)$.
The generator is repeated $k$ times to generate the desired $5^\circ\times k$~degrees rotation of the input image.
When $k>1$ the IDL can still be used to discriminate the intermediate results sampled from $\{B^i\}_{i=1}^{k-1}$.
Models trained using any target rotation are indicated with `A'.

\subsubsection{Learning with a stepwise approach}
\label{sec:stepwise}
Finally we include a model which learns first to rotate objects by a small angle, before learning to propagate the generated images for a larger rotation angle.
On one hand this is inspired on the insight that learning small rotations is easier than larger rotations.
On the other hand, on our experiments which shows that IterGANs --- even with supervised IDLs --- produce artefacts in the intermediate generated images.
This can be overcome if the network first learn to produce high quality small rotations, and then adjust these weights when propagating images for larger rotations.
Therefore we start training the model using $k=1$ for a few epochs, in order to learn a $5^\circ$ rotation. 
We then increase the range of $k$ from 1 to 36 in 3 steps, to increase the degree of rotation from $5^\circ$ to the full $180^\circ$ rotation.
Models trained using this stepwise approach are indicated with `S'.

\section{Experiments}
\label{sec:ex}

\begin{table}
    \centering
    \resizebox{\columnwidth}{!}{
    \begin{tabular}{|cp{.8\columnwidth}|}\hline
		\multicolumn{2}{|c|}{\textbf{Overview of naming of IterGAN models}}\\
		\wip{ig6} &  Using fixed iterations $k=6$ and targets $30^\circ$.\\
		\wip{iga} & Trained using $k$ as control variable, \cf~\autoref{sec:control}\\
		\wip{igk} & Trained with the stepwise approach, \cf~\autoref{sec:stepwise}\\
		\multicolumn{2}{|c|}{\textit{Loss function extensions}}\\
		-M		& trained with the mask objective, \cf~\autoref{eq:L1m}\\
		-U		& trained with \textbf{unsuprevised} IDL\\
		-S		& trained with \textbf{suprevised} IDL\\\hline
    \end{tabular}
    }\\
    \resizebox{\columnwidth}{!}{
    \begin{tabular}{|p{4cm}p{5cm}|}\hline   
        \bf\emph{Generator Loss} & \bf\emph{Discriminator Loss}\\
        {\vspace{0mm}\begin{align*}
                \LL_\Gen  &= H[\Dis_\phi(A, B^k), 1] \\
                            &\quad + \lambda_{L_1} \ L_1 \textrm{\color{MidnightBlue} or } L_1^\textrm{M}(B^k, T)\\
                                &\quad + \lambda_u \ H[\Dis_\mu(B^i), 1] \\
                                &\quad + \lambda_s \ H[\Dis_\nu(A,B^i), 1]
            \end{align*}\vspace{-8mm}}
            &
        {\vspace{-12mm}\begin{align*}
                \LL_\Dis &= H[\Dis_\phi(A, B^k), 0]\\
                                                    &\quad + H[\Dis_\phi(A, T), 1] \\
                                                        &\quad + \lambda_u \ \left( H[\Dis_\mu(B^i), 0] + H[\Dis_\mu(A \lor T), 1] \right)\\
                                                        &\quad + \lambda_s \left( H[\Dis_\nu(A,B^i), 0] + H[\Dis_\nu(A,T^i), 1] \right)
            \end{align*}\vspace{-8mm}}
        \\\hline
    \end{tabular}
    }\vspace{-3mm}
    \caption{Overview of the IterGANs models used in our experiments, including the naming conventions and the generator and discriminator losses.%
    }
    \vspace{-3mm}
    \label{tab:lossoverview}
    \label{tab:modelnaming}
\end{table}

\begin{table*}[t]
    \centering
    {{\newcommand{\mpm}{$\pm$}

	\resizebox{\textwidth}{!}{
    \begin{tabular}{cc|cc|ccccc|cccc}
                       &&    Identity                    &    Projective   & \wip{p65} &    \wip{p30}     &    \wip{ig6}     &    \wip{ig6u}    &    \wip{ig6s}     & \wip{pm30}&    \wip{ig6m}     &    \wip{ig6mu}            &    \wip{ig6ms}     \\\hline%
    $L_1$        &$\downarrow$      & .022\mpm.020                   & .138\mpm.032    & .021\mpm.009          & .014\mpm.009     & .014\mpm.008     & .016\mpm.010     & .016\mpm.010      & .013\mpm.009             & .013\mpm.009      & \textbf{.012\mpm.008}     & \textbf{.012\mpm.008} \\
    $L_1^{\textrm{M}}$ &$\downarrow$ & .298\mpm.154                   & .457\mpm.156    & .299\mpm.092          & .210\mpm.092     & .200\mpm.084     & .239\mpm.099     & .247\mpm.094      & .157\mpm.092             & .162\mpm.060      & \textbf{.147\mpm.055}     & .152\mpm.058          \\
    $D_{\textrm{KL}}$  &$\downarrow$ & \textbf{\emph{.480 \mpm .520}} & 1.407 \mpm .810 & 1.488 \mpm 1.333      & 1.329 \mpm 1.333 & 1.234 \mpm 1.261 & 1.567 \mpm 1.301 & 1.555 \mpm 1.306  & 1.324 \mpm 1.333         & 1.219 \mpm 1.240  & \textbf{1.019 \mpm 1.134} & 1.057 \mpm 1.125      \\
SSIM              &$\uparrow$ & .915 \mpm .068                 & .547 \mpm .082  & .910 \mpm .056        & .938 \mpm .047   & .941 \mpm .045   & .930 \mpm .049   & .931 \mpm .048    & .937 \mpm .052           & .940 \mpm .047    & \textbf{.946 \mpm .043}   & .943 \mpm .046        \\
VIFp               &$\uparrow$ & .446 \mpm .122                 & .154 \mpm .050  & .424 \mpm .098        & .504 \mpm .097   & .513 \mpm .097   & .476 \mpm .096   & .469 \mpm .095    & .492 \mpm .107           & .503 \mpm .102    & \textbf{.522 \mpm .097}   & .517 \mpm .101        \\
    \end{tabular}
    }\vspace{-3mm}}}
    \caption{Comparison of several IterGAN variants and baselines, including supervised and unsupervised intermediate discriminators and training using the \wip{L1M} loss. 
    While the performance differences between some of the models are small, the results are statistical significant for \wip{ig6} over \wip{p30}, and \wip{ig6mu} and \wip{ig6ms} over \wip{pm30}, see text for details.
    \emph{We conclude, that the iterative nature of IterGANs are beneficial for learning object rotation.}
    }
    \label{tab:overview}
\end{table*}

\myparagraph{Dataset}
For most of our experiments we use the Amsterdam Library of Object Images (ALOI)~\citep{geusebroek2005amsterdam} dataset, a set of real images of 1000 household objects, photographed under constrained lighting and from different viewing directions using a turntable setup.
Each object is rotated $360^\circ$ in steps of $5^\circ$ (resulting in 72 images per object), padded and scaled to fit 256x256.

For training we have $28.8$k images from 800 objects (36 pairs of $30^\circ$ rotation per object).
For testing we use two sets,
(i) \textbf{seen objects}: 3.6k images from 100 objects from the training set, yet with different start/end rotations;
(ii) \textbf{unseen objects}: 3.6k images from 100 objects not present during training. 

\myparagraph{Evaluation Measures}
Evaluating the quality of generated images is hard by itself~\citep{wang2002image,salimans16nips}, therefore we use different evaluation measures:

\begin{enumerate}
    \item the pixel-wise $L_1$ loss (also used in~\citep{pix2pix2016});
    \item the object specific mask loss (\wip{L1M} loss, \autoref{eq:L1m}); 
    \item the Kullback-Leibler (KL) Label divergence:
        \vspace{-2mm}\begin{equation}D_{\textrm{KL}}\left(\, p(y|B^k) \, || \, p(y|T) \,\right),\vspace{-2mm}\end{equation}
        to measure the similarity of the label distributions $p(y|\cdot)$, obtained from a pre-trained VGG16, between the generated image $B$ and the target image $T$;
    \item Structural Similarity (SSIM)~\citep{wang2004image}; and
    \item Visual Information Fidelity on the pixel domain (VIFp)~\citep{sheikh2006image}.
\end{enumerate}

The KL-Label measure is inspired on the KL measure used to measure specificity and diversity in~\citet{salimans16nips}, yet in our case the generated image should be realistic and therefore have a similar label distribution to the target image.
Note that for $L_1$, \wip{L1M} and $D_{\textrm{KL}}$ better performing models obtain a lower ($\downarrow$) score, while for SSIM and VIFp better performing models obtain a higher ($\uparrow$) score.

\myparagraph{Training procedure}
For training all models, we followed the training setup of Pix2Pix~\citep{pix2pix2016} as closely as possible.
All models are trained for 20 epochs, alternating between one gradient descent step on $\LL_\Gen$ and one step $\LL_\Dis$.
For $\LL_\Gen$ we maximize $\log(H)$ instead of minimizing $\log(1-H)$, while for $\LL_Dis$ we divide the loss by 2.
As hyperparameters we use $\lambda_{L_1} = 100$ (cf. Pix2Pix) and we set $\lambda_u = \lambda_s = 0.1$.

The final losses used for the generator and discriminator, combined from \autoref{eq:mm-lgen-m1} - \autoref{eq:mm-ldis-m4}, are summarised in \autoref{tab:lossoverview}.
To counteract bad initialisation, each model was trained multiple (3) times, and the best were used for comparison.
Source code, the trained models, and the train and test splits used are available on GitHub\footnote{Code available at: \url{https://github.com/tomaat/itergan}}.

\myparagraph{Models and Baselines}
In the experiments below, we use the following baselines, models and naming conventions:
\begin{itemize}
    \setlength{\parskip}{0pt}
    \setlength{\itemsep}{0pt plus 1pt}
    \item[\textbf{Id}] the identity mapping (B=A);
    \item[\textbf{Proj}] a non-learning projective transformation, which rotates the image plane assuming a pinhole camera to compute point-pairs for the transformation matrix;
    \item[\textbf{P2P}] direct image rotation using Pix2Pix, we compare two variants: \wip{p30} where a $30^\circ$ rotation is learned directly (identical to IterGAN with  k=1), and \wip{p65} where a $5^\circ$ rotation is learned, following the insight that smaller rotations are easier and applied 6 times to obtain the target $30^\circ$ rotations; and
    \item[\textbf{\wip{ig6}}] the proposed IterGAN models using $k=6$, including models with the unsupervised intermediate discriminator (\wip{ig6u}) and the supervised (\wip{ig6s}) variant. We use $k\!=\!6$ for most IG models, since the targets are available at $5^\circ$ intervals. We also compare training with all available rotations \wip{iga} and \emph{stepwise} approach \wip{igk}, see also~\autoref{tab:modelnaming}.
\end{itemize}
For the learning based models we include models trained with the L1 and \wip{L1M} objective (\autoref{eq:L1m}), the latter denoted with \textbf{M}, \eg \wip{ig6mu} for the \wip{ig6} model with the unsupervised intermediate discriminator, trained using the \wip{L1M} loss.

\begin{figure}[t]
    \centering
    \includegraphics[width=.45\textwidth]{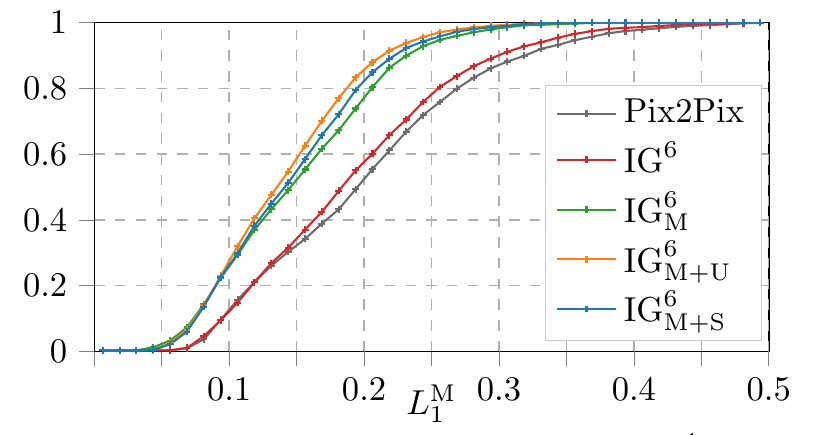}
    \caption{%
        Evaluation of the \wip{L1M}-objective and Intermediate Discriminator Losses, showing the data ($\%$) for a specific loss value.
        The \wip{L1M} objective increase performance significantly, and combined with unsupervised IDL it performs best.
    }    
    \label{fig:idl}
\end{figure}

\vspace{-3mm}
\subsection{IterGANs on ALOI}
\vspace{-2mm}
In the first experiment we compare the baseline methods to several IterGANs variants.
The target is to rotate an input image for $30^\circ$ and we evaluate all measures on the \emph{seen-objects} test set.

\begin{figure*}[p] 
    \centering
    \centerline{\includegraphics[width=1.1\textwidth]{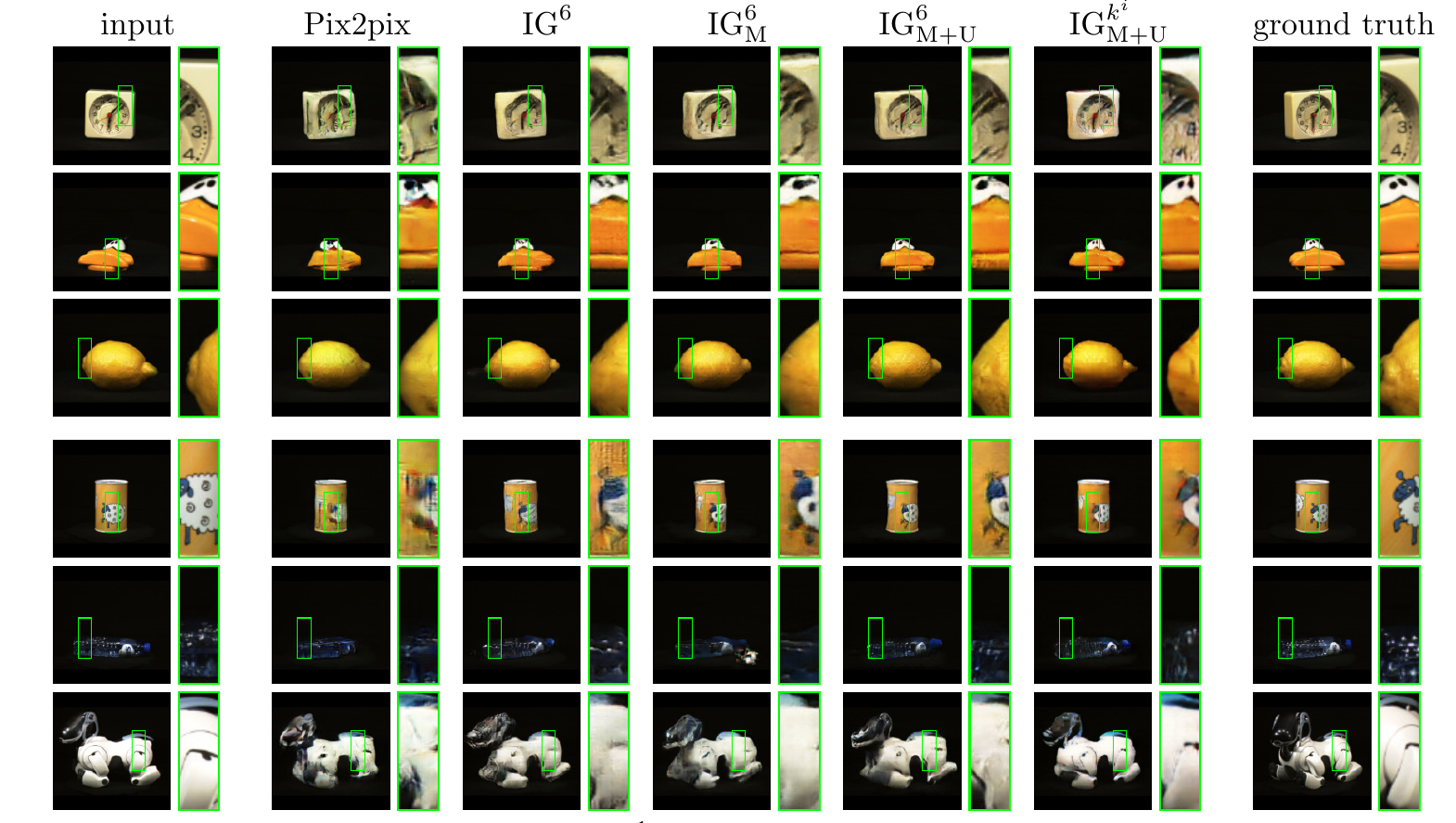}}
    \caption{
        Qualitative comparison of the models (\emph{columns}). The top three rows show a $30^\circ$ rotation of seen objects, and the bottom three for unseen objects.
        The green rectangle is magnified to better compare details.
    }
    \label{fig:examples}
\end{figure*}

The results of this experiment are shown in \autoref{tab:overview}.
From these results, we observe that for evaluation measures $L_1$, \wip{L1M}, SSIM, and VIFp the learning methods outperform the non-learning baselines, while for $D_{\textrm{KL}}$ the identity projection is very strong.
The strong performance of the identity projection for $D_{\textrm{KL}}$ is explained by the VGG16 network, which is (partly) trained to be invariant to object viewpoint, while subtle differences in local image statistics can have a large impact. 
In \autoref{fig:examples} we show some qualitative results of $30^\circ$ rotations of images from the seen objects test set (top three rows). 

In general the IG models improve over \wip{p2p} baselines. Albeit these differences are small, they are significant according to the non-parametric Friedman test where each image-pair judges different models. 
The test finds models which are significantly ranked higher or lower than the others, \eg \wip{ig6} is significantly better than \wip{p30} with a $p \leq 0.01$ based on the \wip{L1M} metric, and the same holds for \wip{ig6mu} and \wip{ig6ms} over \wip{pm30}.

\myparagraph{Object mask objective (\wip{L1M})}
Learning with the \wip{L1M} objective shows a clear increase in performance of any model and any evaluation measure, see \autoref{tab:overview}.
This holds especially for the \wip{L1M} and  $D_{\textrm{KL}}$ evaluation measures. 
Probably because the learner is now informed about the important region of the target image, which yields higher quality of the final generated image.

\begin{figure*}[p] %
    \centering
    \centerline{\hspace{-15pt}\includegraphics[width=1.1\textwidth]{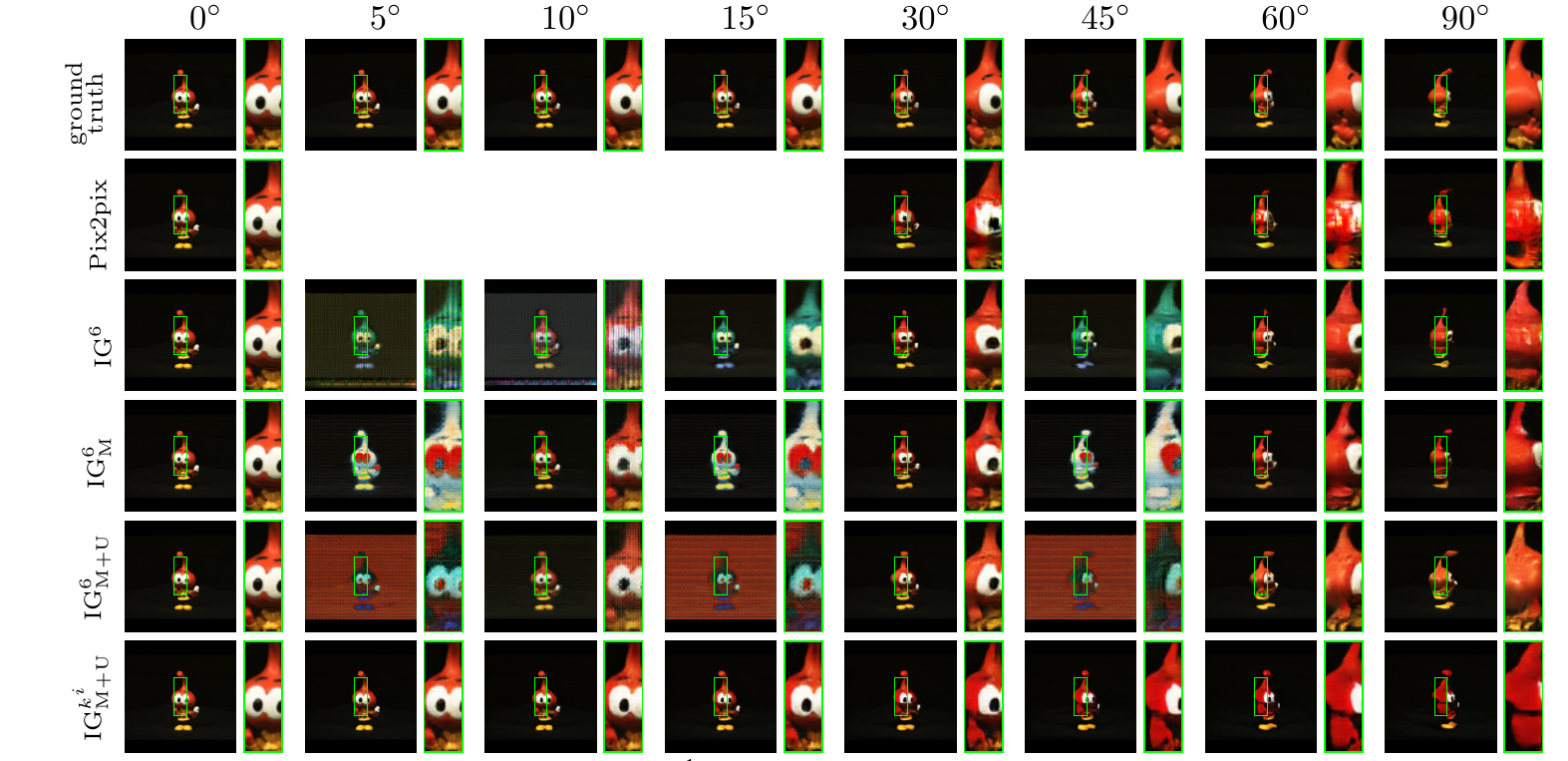}}
    \caption{
        Inter- and extrapolation of the different models (\emph{rows}) with the ground truth at top.
        The columns show the rotation from the input.        
        The IterGANs show more realistic generated images for a wide range of angles.
        \emph{(Best viewed in colour, zoom in for details)}
    }
    \label{fig:ex-between4}
\end{figure*}

\begin{figure*}[t] %
    \centering
    \centerline{\hspace{-15pt}\includegraphics[width=.8\textwidth]{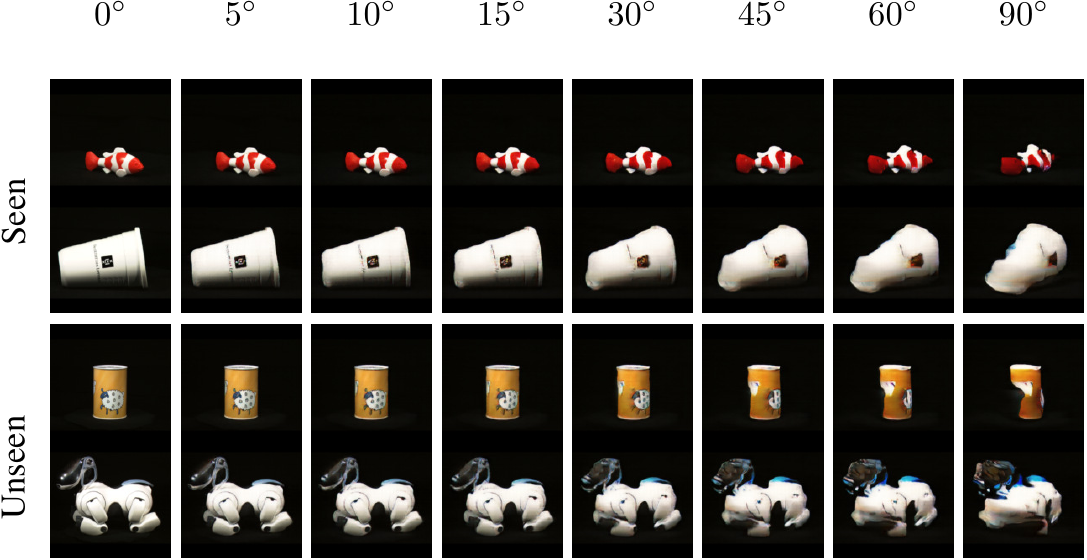}}
    \caption{
    	Illustration of generated rotated objects from \wip{igkmu} for both \emph{seen} and \emph{unseen} objects for different rotation angles.
        \emph{(Best viewed in colour)}
    }
    \label{fig:ex-moreobjects}
\end{figure*}

\myparagraph{Intermediate Discriminator Losses}
Here we look in more detail at the performance when the Intermediate Discriminator Losses (IDL) are added.
The results from \autoref{tab:overview} are detailed in \autoref{fig:idl}, where we show the cumulative data percentage for a given loss, \eg for the \wip{ig6} about 60\% of all images have a loss $<.2$, while for \wip{ig6mu} that is about 85\%.
From \autoref{tab:overview}, we observe that IDL is only beneficial when combined with the \wip{L1M}-objective. 
Apparently there is an interplay between the objective and the generator, that helps in getting a strong enough discriminator to guide the generator. 
From the detailed figure (\autoref{fig:idl}), we observe that IDL is almost only beneficial in the middle regime, for the loss around $.2$, the IDL models obtain ~10\% more examples with such a loss.

Finally from the fact that unsupervised IDL outperforms supervised IDL, see~\autoref{tab:overview} and~\autoref{fig:idl}, we conclude that the models either already perform at their best (which is unlikely given the artefacts), or that the extra ground-truth data is underused by the current training paradigm.
This leaves room for improvement by using a different discriminator in supervised IDL.

\begin{figure}[t]
    \centering
    \includegraphics[width=\columnwidth]{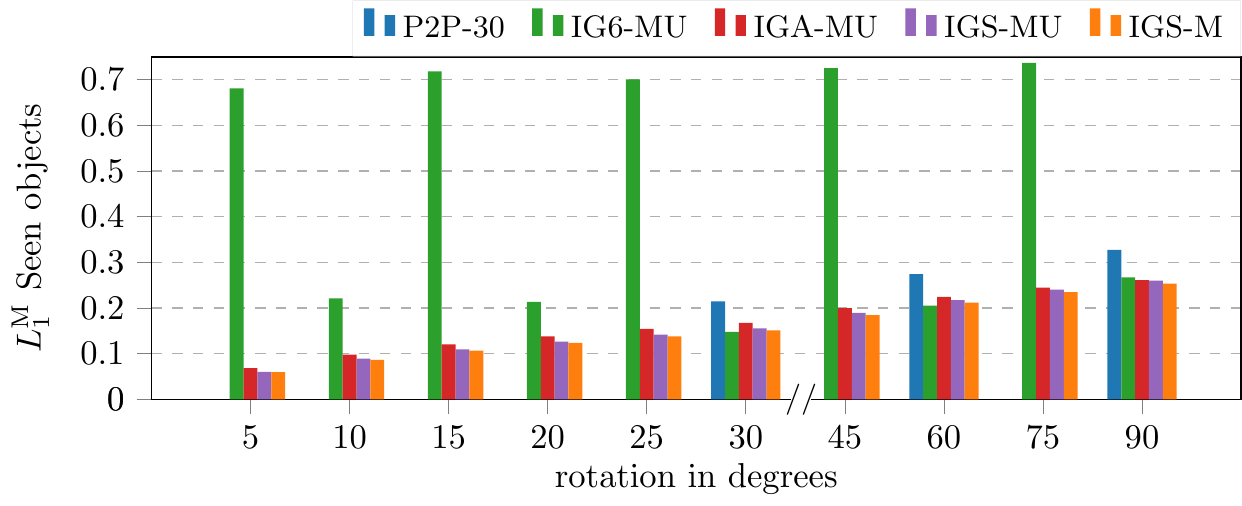}
    \caption{
            Showing the performance versus the target rotation. 
        Note that \wip{p30} only rotates in steps of $30^\circ$, missing most angles.
        Smaller rotations are easier than large rotations, and iterative training performs best for almost any rotation.        
    }
    \label{fig:rot}
\end{figure}

\myparagraph{Stepwise learning: Control object rotation}
Since IterGANs generate the final image in steps, each step could be interpreted as a partial rotation.
In this experiment we use the number of iterations $k$ to control the amount of rotation, \ie by varying $k$ we inter- and extrapolate the generator to generate different angles.
Note that \wip{p30} can only rotate in steps of $30^\circ$ and that the \wip{ig6mu} has no guarantee to rotate for $5^\circ$ per step, it has just been trained on performing $30^\circ$ rotations in 6 steps.
We include three additional training strategies, each using different target rotations and losses:%
\vspace{-2mm}
\begin{enumerate}
    \setlength{\parskip}{0pt}
    \setlength{\itemsep}{0pt plus 1pt}
    \item \wip{igamu}: sampling $k$ from the full range for each epoch;
    \item \wip{igkm}: following the stepwise learning (\autoref{sec:stepwise}); and
    \item \wip{igkmu} the same as above, with the unsupervised IDL.
\end{enumerate}
In~\autoref{fig:rot} we show the \wip{L1M} performance of several models for different angles, thus varying values of target $k$ at evaluation time.
We also provide qualitative results for different models, see~\autoref{fig:ex-between4}, and for different objects (both seen and unseen) for \wip{igkmu}, see ~\autoref{fig:ex-moreobjects}.
From the stepwise learning variants, the graph shows that training with incrementing $k$ (\wip{igk}) outperforms training with all values for $k$ from the start (\wip{iga}).
This indicates that learning small rotations first helps the training process, it also indicates that IDL is of little beneficial value when such an incremental learning approach is used. 

\begin{table}[t]
    \centering
    {\footnotesize
    \begin{tabular}{l|c|c}
                & \wip{L1M} seen & \wip{L1M} unseen \\\hline
Identity        & $.298\pm.154$      & $.295\pm.139$ \\\hline
\wip{p2p}       & $.210\pm.092$      & $.256\pm.100$ \\
\wip{ig6}       & $.200\pm.084$      & $.249\pm.100$ \\\hline
\wip{ig6m}      & $.162\pm.060$      & $.252\pm.094$ \\
\wip{ig6mu}     & $\bf.147\pm.055$   & $.231\pm.094$ \\
\wip{ig6ms}     & $.152\pm.058$      & $.232\pm.092$ \\\hline
\wip{igkm}     & $.155\pm.066$      & $\bf.167\pm.073$ \\
\wip{igkmu}    & $.151\pm.063$      & $\bf.167\pm.071$
\end{tabular}
    }
    \caption{
            Comparison on \emph{seen} and \emph{unseen} test sets.
        The learning based models loose $0.5\!-\!1$ on the \wip{L1M} evaluation metric on the \emph{unseen} data.
        Surprisingly the stepwise \wip{igk} models perform almost equally on the \emph{seen} and \emph{unseen} objects.
    }
    \label{tab:unseen}  
\end{table}

\myparagraph{Discussion on quality of intermediate images}
A clear phenomenon when using \wip{ig6} IterGANs are the artefacts introduced in the intermediate generated images, see row 3-5 in ~\autoref{fig:ex-between4}, which disappear at the final iteration.
The most obvious artefact is the red background added to every other image generated by \wip{ig6mu}.
This results in a phase-two periodic pattern in~\autoref{fig:rot}, where \wip{ig6mu} is either one of the best scoring models (for 30, 60, and 90), or the worst scoring models (\eg for 5, 15, and 25). 

This phenomenon shows that even with IDL, it is hard to train the generator to produce both good final images and high quality intermediate images.
We have evaluated performance of a $30^\circ$ rotation as a function of the rotation of the object on the input image, the standard deviation is low ($0.012 / 0.013$ on $L_1^{\textrm{M}}$).
Based on this evaluation we conclude that the color tint is not caused by the angle of the object on the input image, it is caused by suboptimal parameters in the network
We believe this is due to the complex parameter space, where the set of parameters which satisfies both objectives is smaller than the set of parameters which only produce good final images.

We have investigated this by learning multiple times with the same settings and we observed that this pattern already occurs at an early stage, \eg the red background is visible after a few epochs and the model does not escape from this local maximum to produce higher quality intermediate images.
One way to overcome these artefacts is by first learning for $5^\circ$ rotations before propagating generated images for multiple iterations, the strategy we follow in our stepwise approach (\wip{igk}).

The discriminator in the unsupervised IDL model has to tell apart real images ($A$ or $T$) from generated images ($B^k$), over a large set of objects and colors. 
For this discriminator an image with different colouring, yet consistent local structure, might look real. Another object can have this color pattern.
This could be partly caused by our optimisation strategy, where for every input and output pair, we sample a \emph{single} intermediate image for IDL.
It would be interesting to use all generated images and then enforce a consistency requirement over this sequence.

\begin{table}[t]
	\begin{tabular}{l|ccccc}
			& GT & P2P-M30 & IG6-M & IG6-MU & \emph{Avg Score}\\\hline
	GT 		&  0 		&17 	& 18 	& 15		& 0.74\\
	P2P-M30  & -17 	& 0  	& -7 	& -3 		& -0.13\\
	IG6-M 	&-18 	& 7 	& 0 	& -14	& -0.1\\	
	IG6-MU 	&-15 	& 3	&14 	&0		&0.014\\		
	\end{tabular}
	\caption{Human evaluation of different models. The human act as judge of a pair of images, and selects one as best. The winning image obtains a +1 score, the other -1. For the average score we summed the scores per model
and divided by the number of times the model was shown. The ground-truth image is selected $87\%$ of the time as best, meaning that in the remaining $13\%$ a generated image is deemed more realistic.}
	\label{tab:human}
\end{table}

\subsection{Human Evaluation}
Given that none of the evaluation metrics will be able to properly address image quality, we have performed a (small) user study.
Users would see an input image, and the corresponding output images of two models, or the ground-truth.
The uses were asked to indicate which output image they found the best of the two considering the input image. 
We have compared ground-truth, \wip{pm30}, \wip{ig6m}, \wip{ig6mu}, and in total 342 image pairs are judged by in total about 10 people.

The results are shown in~\autoref{tab:human}, where we sum the scores of the human judges, $+1$ for the image which is judged as best by the human, and $-1$ for the other image. 
When a ground-truth image is shown, it is selected in about $87\%$ of the time (to transform our score average to a selection percentage we use $\tfrac{1}{2}(1+S)$).
Which indicates that in about $13\%$ of the cases the generated image is of such high quality it is selected above the ground-truth image.
For the other models, the results are closer together, but indicate that \wip{ig6mu} is selected more often than \wip{ig6m}, which is selected more often than \wip{pm30}.
This user study confirms our experiments, that the iterative nature of IterGANs helps to generate to more realistic  images.

\vspace{-5pt}
\subsection{Unseen objects}
In this experiment we use the test set with the \emph{unseen} objects, to compare the generalisation of the different models.
The results are shown in \autoref{tab:unseen}.
We observe, as is expected, that the performance on the seen objects is in general better than the unseen objects, and that training for the \wip{L1M} metric does not improve the performance over the unseen objects that much.
The models trained with stepwise learning (\wip{igkm} and \wip{igkmu}), however, generalise extremely well to never seen objects, performing better than \wip{p2p} and \wip{ig6} on seen data. 
In \autoref{fig:examples} (\emph{bottom three rows}) unseen objects are shown, note how the sheep-shape on the mug is created, while that mug was not seen during training. 
In conclusion, our models learn depth cues and texture patterns, to rotate even never seen objects. 

\begin{figure*}[t]
        \centering
        \begin{subfigure}[b]{5.5cm}
        		\includegraphics[height=5cm]{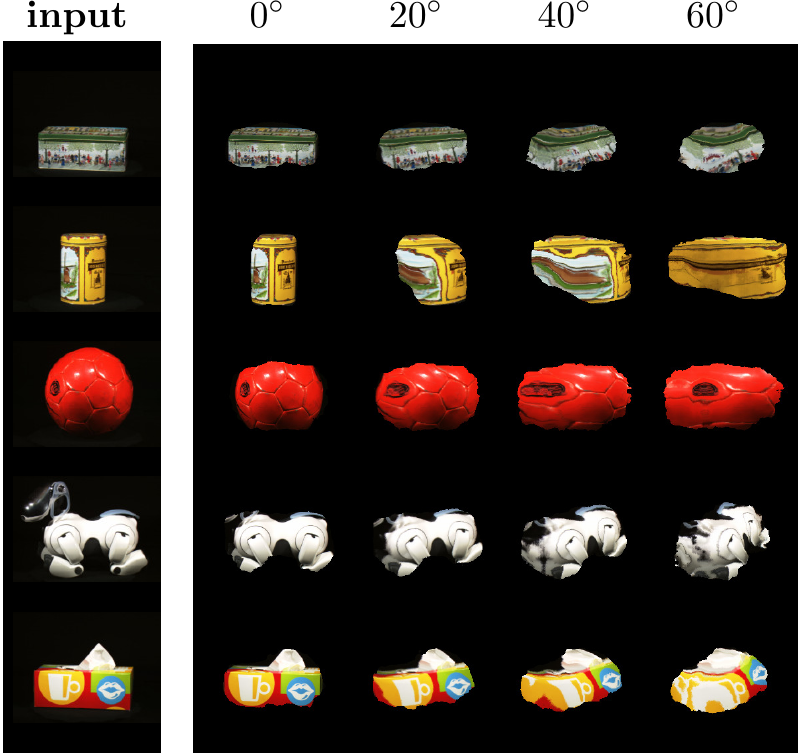}
		\caption{Appearance Flow~\citep{zhou16eccv}}
    	\end{subfigure}
        \begin{subfigure}[b]{3.25cm}
        		\includegraphics[height=5cm,trim={3.5cm 0 0 0},clip]{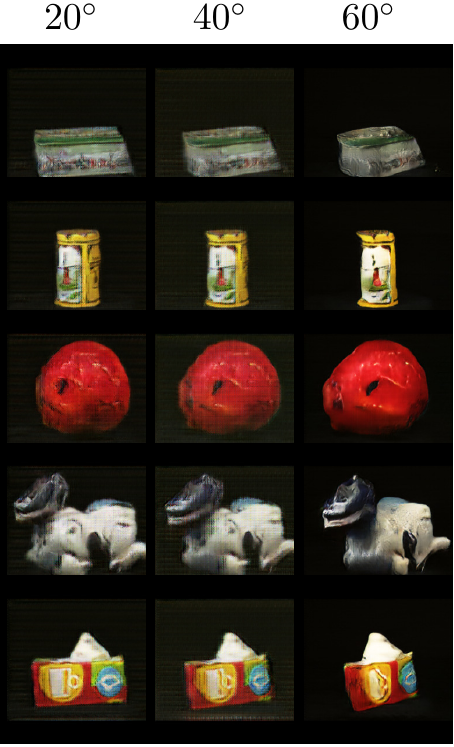}       
		\caption{\wip{ig6mu}}
    	\end{subfigure}	
        \begin{subfigure}[b]{3cm}
        		\includegraphics[height=5cm,trim={3.5cm 0 0 0},clip]{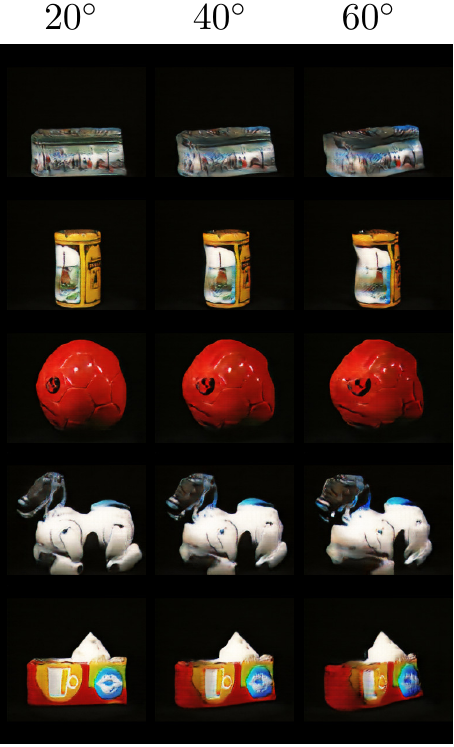}       
		\caption{\wip{igkmu}}
    	\end{subfigure}
	\vspace{-3mm}
        \caption{Illustrative examples of ALOI objects rotated with Appearance Flow~\citep{zhou16eccv}, compared to our proposed \wip{ig6mu} and \wip{igkmu} models.
        Appearance Flow warps all images into a sedan style shape, already for the $0^\circ$ rotation, while the colours (not texture) remain realistic.
        In contrast our models propagate colours less convincingly, yet the original shape is better preserved.
        }
        \label{fig:afunseen}
\end{figure*}

\myparagraph{Comparison to Appearance Flow}
In this section we compare IterGANs to the Appearance Flow (AF) method of~\citet{zhou16eccv}.
AF is specifically trained on cars, from CAD renders, disentangling the flow and the appearance of the rotations.
We use the pre-trained AF model on our \emph{unseen} test set, see \autoref{fig:afunseen} for some qualitative results and compare to the \wip{igkm} model.

The results are presented in \autoref{tab:afcomparison}, where we show the performance of AF and \wip{igkm} for $R=\{20,40,60\}$ rotations, following the AF settings.
For any of the rotations the \wip{igk} model outperforms AF by a large margin, making the explicit control of IterGANs an alternative for the implicit control vector in AF.

\begin{figure*}[t]
  \centering
  \includegraphics[width=.7\textwidth]{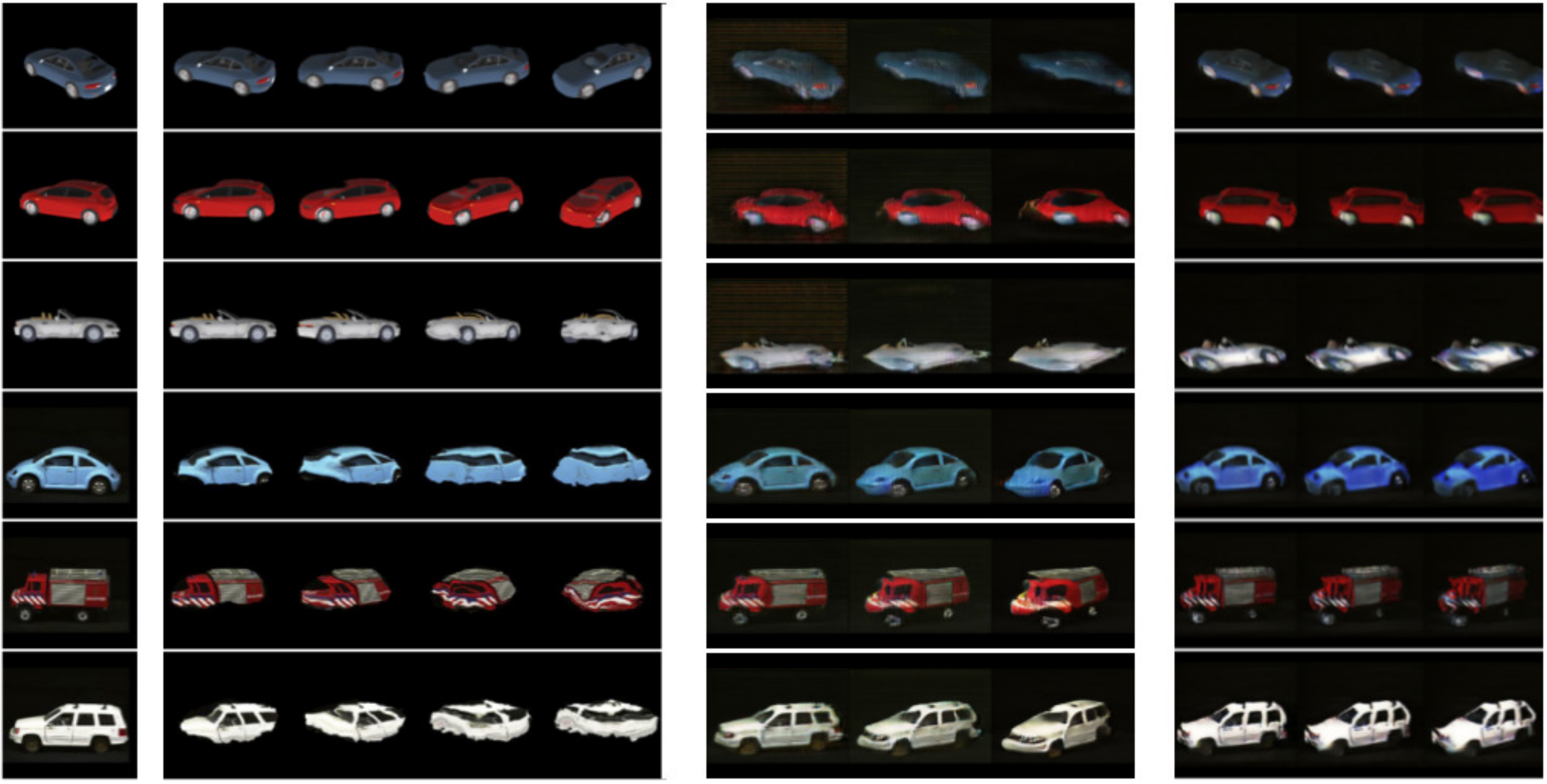}
  \caption{
    Illustration of rotated cars from Appearance Flow~\citep{zhou16eccv}(\emph{left}) and our proposed \wip{ig6mu} (\emph{middle}) and \wip{igkmu} (\emph{right}) models.
    We show three cars from the AF dataset (\emph{top}) and three from ALOI (\emph{bottom}). Note \wip{ig6mu} is only trained on $30^\circ$ rotations, while the shown rotations are 20/40/60 degree.
  }
  \label{fig:appflow}
\end{figure*}

In~\autoref{fig:appflow} we show a qualitative comparison to the AF, using only cars from ALOI and their data set.
From the figure we conclude that both methods are specific about the expected transformation/rotation, and have difficulty to generalise well to the other dataset.
AF is (logically) more specific about the object, it warps every object into a sedan, but transfer realistic/correct colours.
IterGANs, on the other hand, are agnostic towards the kind of object to rotate, introduce some color changes, and are specific on the turntable setup.

\begin{table}[t]
    \centering
    {%
    \small
    \begin{tabular}{|l|cccc|}\hline
    	& L1 ($\downarrow$) & L1M  ($\downarrow$) & VIFp ($\uparrow$)  & SSIM ($\uparrow$)\\\hline\hline	
	&   \multicolumn{4}{c|}{$20^\circ$ rotation} \\\cline{2-5}		
	AF 			& .023 & .187 & .102 & .402\\
	\wip{ig6mu} 	& .030 & .249 & .228 & .781\\
	\wip{igkm} 	&\bf .012 &\bf .125 &\bf .322 &\bf .834\\\hline\hline
	&   \multicolumn{4}{c|}{$40^\circ$ rotation} \\\cline{2-5}		   
	AF 			& .030 & .218 & .202 & .392\\
	\wip{ig6mu} 	& .049 & .323 &\bf .435 & 636\\	
	\wip{igkm} 	&\bf .015 &\bf .156 & .302 &\bf .818\\\hline\hline	
	&   \multicolumn{4}{c|}{$60^\circ$ rotation} \\\cline{2-5}		
	AF 			& .034 & .227 & .066 & .386 \\	
	\wip{ig6mu} 	& .025 & .308 & \bf .390 &\bf .895\\	
	\wip{igkm} 	&\bf .016 & \bf .179 & .291 & .808\\\hline	
    \end{tabular}

    }
    \caption{Comparison of Appearance Flow \citep{zhou16eccv} and the proposed IterGANs on the unseen test set of ALOI.
    Note that AF is originally trained on a dataset of synthethic cars, while \wip{ig6mu} only on rotations of $30^\circ$.
    IterGANs obtain higher performance on any rotation and any evaluation measure.
    }
    \label{tab:afcomparison}
\end{table}

\myparagraph{IterGANs for Data Augmentation}
In this section we explore using IterGANs as a means of data augmentation in a classification system.
Unfortunately, the ALOI dataset is not suitable for classification, per class it only contains a single object (with photos taken from various viewpoints). 
We resort to the Office-Home dataset~\citep{venkateswara2017Deep} with real object centric images, which somewhat resemble the ALOI data used for training IterGANs. 
We evaluate data augmentation in a domain transfer \& classification experiment, where IterGANs are used to generate novel views of images from a novel domain.

The product domain of the dataset contains 4439 images from 65 categories, which we use as follows:
for each class, 9 images are used for testing (585 in total), 5 images for validation (325 in total), and the remaining 3529 images for training. 
The minimum number of example images for a class is 38 (resulting in 24 training images for that class). 
We have automatically changed the white background for a black background to resemble the ALOI images.

\begin{figure}[t]
	\centering
        \includegraphics[width=.8\columnwidth]{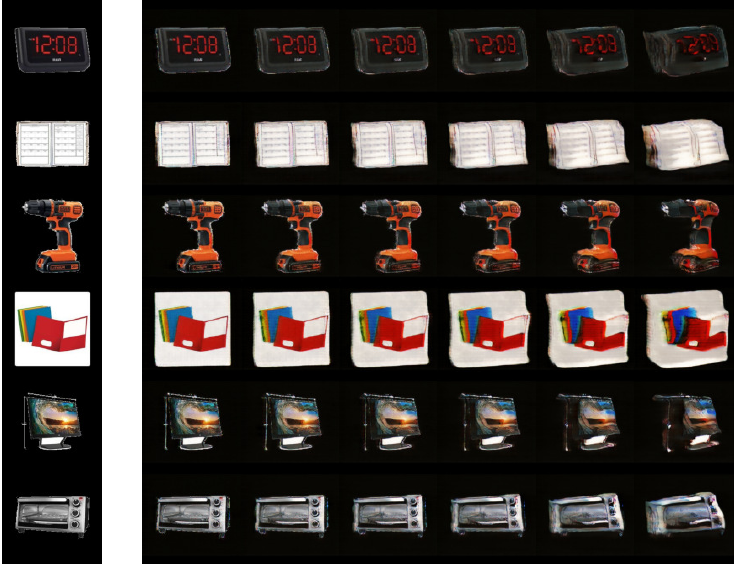}
        \caption{Examples of Product dataset with rotated versions}
        \label{fig:pproduct}
\end{figure}

On this dataset we have trained and evaluated the following: %
\vspace{-1mm}
\begin{itemize}
    \setlength{\parskip}{0pt}
    \setlength{\itemsep}{0pt plus 1pt}
    \item Train on the original dataset (3529 images); or
    \item Train on an augmented training set, either \emph{rot3} where IterGANs are used to generate 3 images: $5^\circ$, $10^\circ$, and $15^\circ$ rotations, or \emph{rot6}, where 6 images are generated (also  $20^\circ$,$30^\circ$, and $40^\circ$), see~\autoref{fig:pproduct} for some illustrative examples.
    \item Evaluate using the original image only, or an average prediction of the original plus generated rotated versions.
\end{itemize}
For these experiments we have used fixed image representations (from VGG or ResNet-101) and used the validation set for the learning rate and the number of epochs, using the performance of evaluating the original images only.

The results are shown in~\autoref{tab:pproduct}.
We observe that using IterGANs for data augmentation can always be beneficial over using only the original data at test and train time.
The best performance is obtained when trained on the original data and evaluated using a small rotated version of the object.

\newcommand{\rotatedcell}[1]{%
\multirow{3}{*}{\rotatebox[origin=c]{90}{\parbox[c]{12mm}{#1}}}
}
\begin{table}[t]
    \centering
    {\footnotesize
    \begin{tabular}{ll|c@{\hspace{2mm}}c@{\hspace{2mm}}c@{\hspace{2mm}}c@{\hspace{2mm}}c@{\hspace{2mm}}c@{\hspace{2mm}}c}
                                            &           & org\phantom{50}   & +R5\phantom{0} & +R10 & +R15 & +R20 & +R30 & +R45\\\hline
\rotatedcell{VGG16}         & org   & 76.7 & \textbf{\color{RoyalBlue}{79.0}} & 77.6 & 76.9 & 74.9 & 73.7 & 72.7\\
                    & rot3  & 77.4 & \textbf{78.3} & 76.6 & 75.9 & 75.4 & 74.4 & 73.2\\
                    & rot6  & \textbf{76.2} & \textbf{76.2} & \textbf{76.2} & 76.1 & 74.5 & 73.0 & 73.0\\\hline 
\rotatedcell{ResNet101} & org   & 70.8 & \textbf{\color{RoyalBlue}{75.9}} & \textbf{\color{RoyalBlue}{75.9}} & 74.5 & 73.7 & 71.3 & 68.6\\
                    & rot3  & 72.5 & \textbf{74.9} & 74.7 & 74.4 & 73.2 & 72.7 & 71.6\\
                    & rot6  & 66.7 & 70.9 & 70.6 & 71.8 & 72.3 & \textbf{72.5} & 71.5\\\hline
    \end{tabular}
    }
    \caption{Classification accuracy for different variants of using IterGANs for data augmentation. 
    Using rotation augmentation is always helpful at test time.
    }
    \label{tab:pproduct}
\end{table}

\subsection{Camera rotation on VKITTI}
In our final experiment, we explore the problem of camera rotation: \emph{generate a scene from a different camera viewpoint}.
For this task we use the Virtual-KITTI dataset~\citep{Gaidon:Virtual:CVPR2016}, and the task is to generate a $30^\circ$ rotated side-view from the main camera.
We train on 4 sequences, using images of $728\times256$ pixels, since the model is fully convolutional, the number of parameters remains the same. 
This data set is smaller than ALOI, therefore we train for 50 epochs, since the data does not contain object masks, nor intermediate images, we can not train using the \wip{L1M} objective or the supervised IDL (\wip{igs}) models. 

\begin{figure}[t] %
    \centering
    \let\temp\tabcolsep
    \setlength{\tabcolsep}{0em}
    \begin{tabular}{ccc}
    \rotatebox{90}{~\hspace{.1em} output \hspace{5.5em} intermediate images \hspace{7em} input} &
    \includegraphics[width=.48\linewidth]{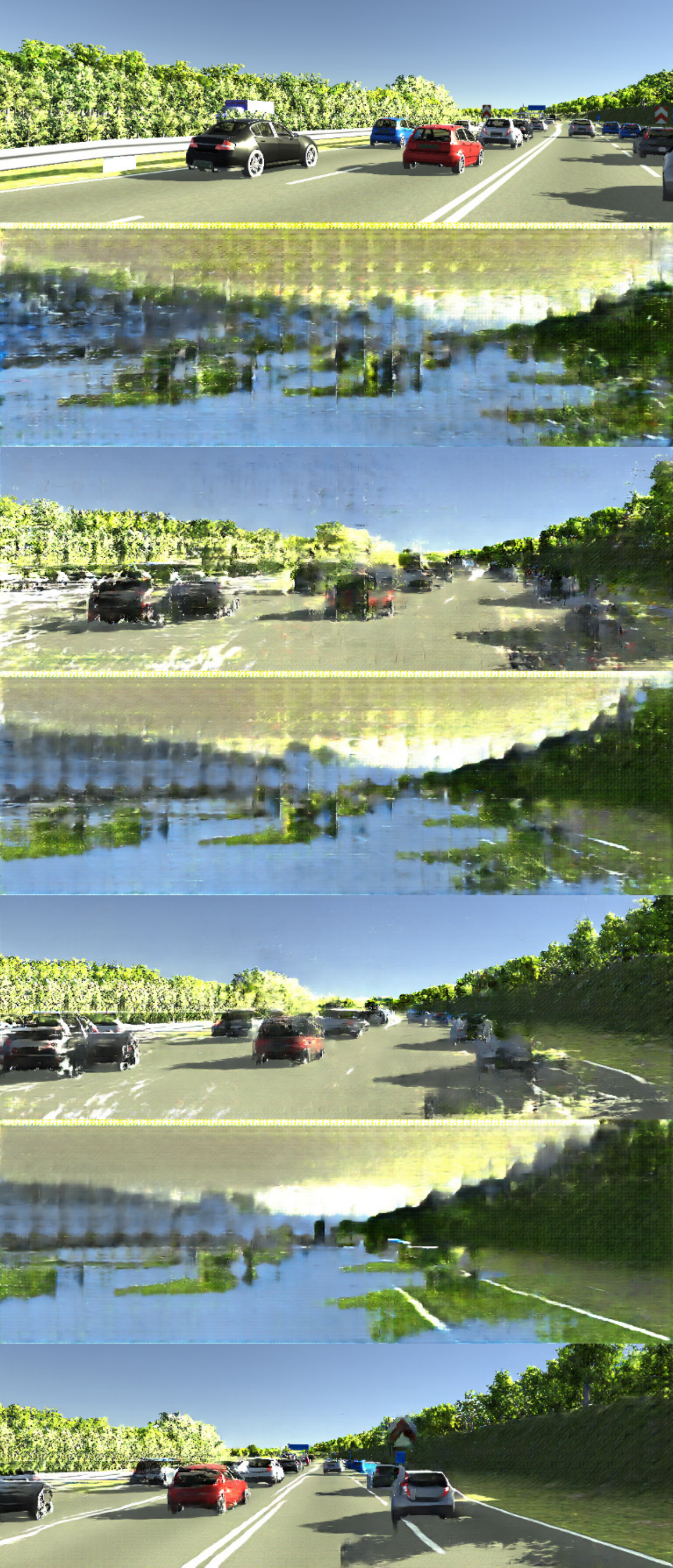} &
    \includegraphics[width=.48\linewidth]{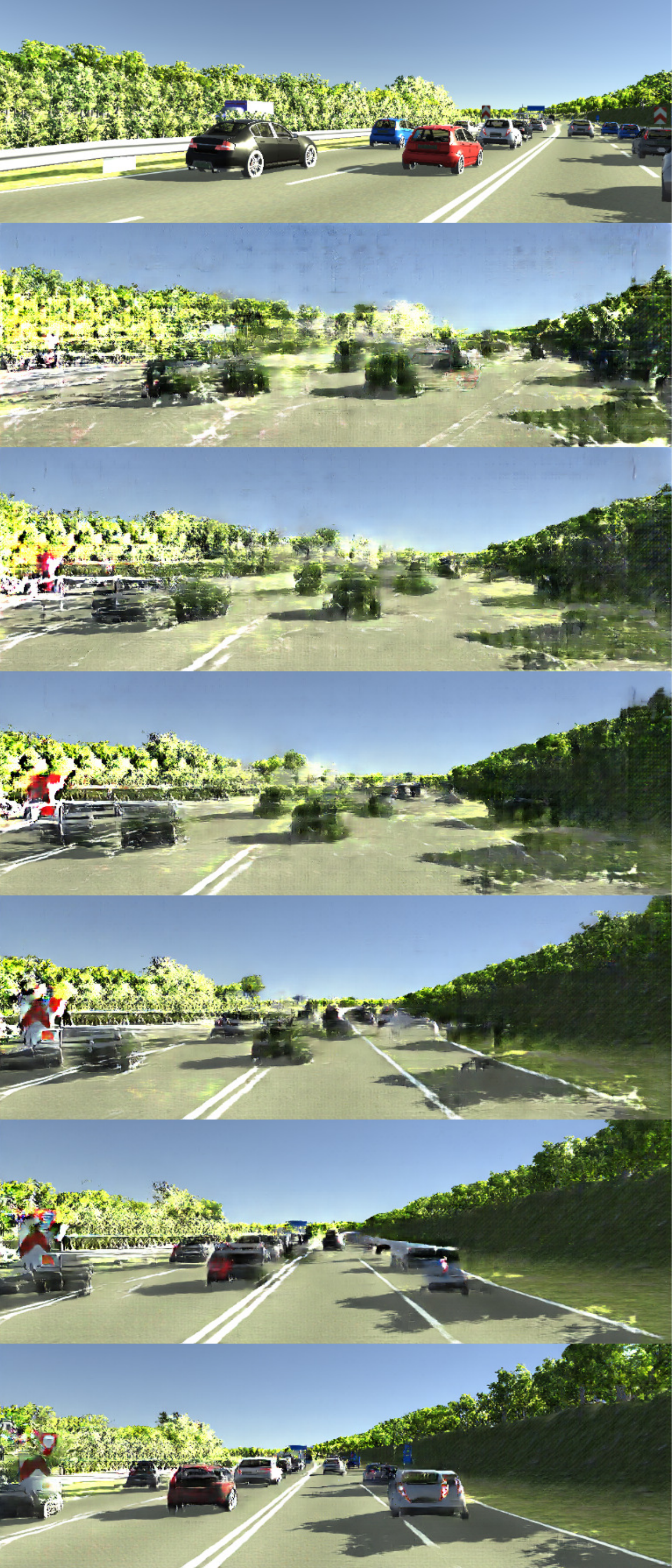}\\[-1mm]
    \rotatebox{90}{~\hspace{.1em} target}&
    \includegraphics[width=.48\linewidth]{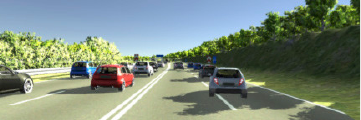} &
    \includegraphics[width=.48\linewidth]{vkitti_int_target}\\[-2mm]
    & \wip{ig6} & \wip{ig6u}\\[-4.5mm]
    \end{tabular}
    \setlength{\tabcolsep}{\temp}
    \caption{
        Illustration of intermediate generated images from the VKITTI dataset. The \wip{ig6} model seems to add/remove details, while the
    \wip{ig6u} model iteratively rotates and add details until the final output image.
    }
    \label{fig:vkitti-between}
\end{figure}

We train using \wip{ig6} and \wip{ig6u}, and show intermediate generated images in \autoref{fig:vkitti-between}.
The \wip{ig6} model seem to alternate between adding details and loosing details, it is almost surprising that from this final-last image such a realistic output image is synthesised.
The \wip{ig6u} model, on the other hand, seems to iteratively add details and rotate the image, yet with a big shift in rotation in the first image and then adding details. The final image has some different colourings than the target image. 

\begin{figure}[t]
    \centering
    \includegraphics[width=.45\textwidth]{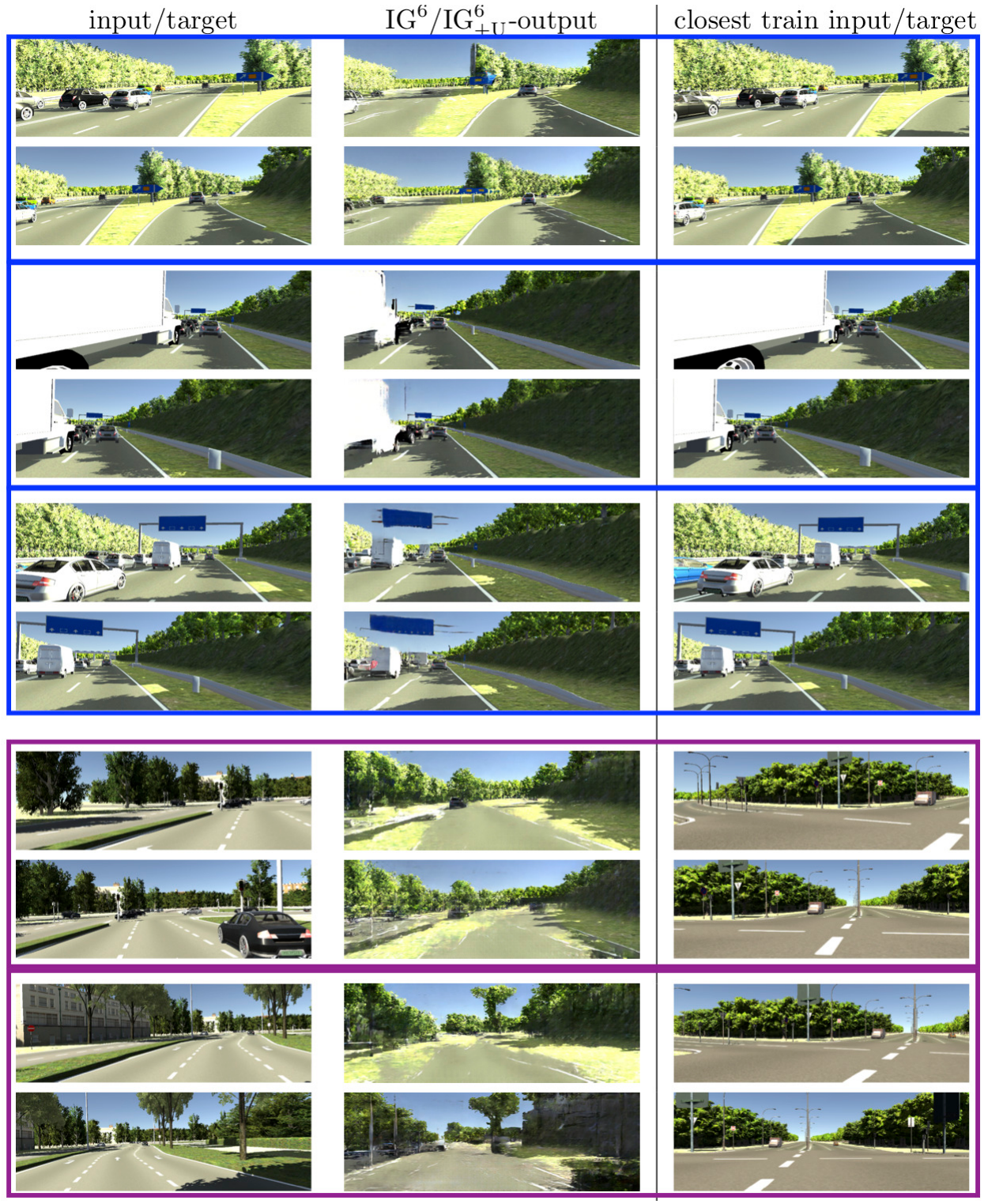}
    \caption{
	Qualitative results on VKITTI dataset for test images from seen \emph{(top)} and unseen \emph{(bottom)} sequences. 
	For each showing input/target images, generated results from \wip{ig6} and \wip{ig6u}, and nearest training images ($L_2$ distance).	
	The examples from unseen sequences show many artefacts, the difficulty of the problem is highlighted by the dissimilarity with the nearest training images. 
    }
    \label{fig:vkitti}
\end{figure}

In \autoref{tab:vkscore} we show the quantitative results of three models: \wip{p2p}, \wip{ig6}, \wip{ig6u}.
The performance is rather similar, yet the \wip{ig6u} model performs better than the other two models.
In \autoref{fig:vkitti} we show qualitative results, of input/output pairs with the generated output images and the closest training example (based on L2 distance).
The quality of the generated images on the test parts of the seen sequences are very good, yet on the unseen sequences are not convincing. 
This might be due to overfitting on the training set, yet the synthesised images are different from the closest train input and looking very realistic.

\begin{table}[t]
    \centering
    {\footnotesize
     \begin{tabular}{lcc}
                & seen & unseen \\ \hline
            Pix2pix    & $0.077\pm0.017$ & $0.216\pm0.019$ \\
            \wip{ig6}  & $0.073\pm0.015$ & $0.209\pm0.020$ \\
            \wip{ig6u} & $0.073\pm0.015$ & $0.201\pm0.024$\\[-3mm]
    \end{tabular}
    }
    \caption{Comparison of models on VKITTI dataset, using $L_1$ evaluation metric.
    }
    \label{tab:vkscore}
    \vspace{-3mm}    
\end{table}

\section{Conclusion \& Outlook}
\label{sec:conc}
In this paper we have introduced IterGANs, a GAN model which iteratively transforms an image into a target image whereby the generator has to learn only small transformations.
IterGANs are in part inspired on the \citet{shepard1971mental} mental rotation experiment\footnote{In \autoref{fig:mental} the left objects are not only rotated, but also mirrored.}, which indicates that learning small rotations is easier than larger rotations.
The iterative nature of IterGANs also allows for additional discriminators in the objective function, either supervised or unsupervised, on the intermediate images.
These discriminators help to overcome some artefacts, and lead to better final synthesised images.

Our experiments have shown that IterGANs outperform a direct transformation GAN (\wip{p2p} model).
Surprisingly the unsupervised intermediate discriminator loss is on par with the supervised counterpart, indicating that the additional supervision is not used optimally.
A more extensive exploration of possible intermediate loss functions, \eg exploiting the sequence of the generated images rather than sampling a single intermediate image is left for future work.

We have then explored using the additional supervision in a stepwise training approach: in the first few epochs the model learns only small rotations of objects, before learning larger rotations. 
We have also used the number of iterations in the network as an explicit control signal.
This helps to guide the learning process and produce more realistic images, especially for \emph{never seen} objects.

Future research could investigate incorporating the explicit control which is offered by the number of iterations used with IterGANs into an implicit control vector.
For example the iterative process could also output a (residual) control vector, and the generated image has the target rotation when the control vector is all zero. 
While our model only learns right rotations of an object, such a more flexible model could be trained with \emph{cycle awareness}~\cite{Zhu_2017_ICCV}.
This exploits the insight that a realistic generated image from an input with a target transformation could also be used as input with the inverse transformation to obtain the input image back.
Such an explicit-implicit controlled IterGANs could allow for full 3D transformation of both objects and scenes.

\section*{Acknowledgements}
We thank all reviewers for the extensive feedback and suggestions for improving our manuscript.
This research was supported in part by the NWO VENI What\&Where project.

\bibliographystyle{model2-names}
\bibliography{bib}

\end{document}